\documentclass[journal]{IEEEtran}

\usepackage[utf8]{inputenc}
\usepackage{siunitx}
\usepackage[colorlinks=true, allcolors=blue]{hyperref}
\usepackage{amsmath}
\usepackage{amsfonts}
\usepackage{soul} %ONLY FOR THE DRAFT VERSION, REMOVE AFTERWARDS
\usepackage{subcaption}
\usepackage{stfloats}
\usepackage{booktabs}
\usepackage{enumitem}

\usepackage{tikz,xcolor}

\definecolor{lime}{HTML}{A6CE39}
\DeclareRobustCommand{\orcidicon}
{
    \begin{tikzpicture}
    \draw[lime, fill=lime] (0,0) circle [radius=0.16] 
    node[white] {{\fontfamily{qag}\selectfont \tiny ID}};    \draw[white, fill=white] (-0.0625,0.095) circle [radius=0.007];    
    \end{tikzpicture}
    \hspace{0mm}}
\foreach \x in {A, ..., Z}{%
    \expandafter\xdef\csname orcid\x\endcsname{\noexpand\href{https://orcid.org/\csname orcidauthor\x\endcsname}{\noexpand\orcidicon}}
}

\setlist[enumerate]{itemsep = 0pt, parsep = 0pt, topsep = 0pt} % enumerate spacing setting
\setlist[itemize]{itemsep = 0pt, parsep = 0pt, topsep = 0pt} % itemize spacing setting

\begin{document}

\title{Development of Occupancy Prediction Algorithm for Underground Parking Lots}

% https://www.programmersought.com/article/12636730965/
\author{Shijie Wang \vspace{-20pt}
}

% The paper headers
\markboth{Journal of \LaTeX\ Class Files,~Vol.
% ~14, No.~8, August~2015
}%
{Shell \MakeLowercase{\textit{et al.}}: Bare Demo of IEEEtran.cls for IEEE Journals}

% make the title area
\maketitle

% As a general rule, do not put math, special symbols or citations
% in the abstract or keywords.
\begin{abstract}
Bird's Eye View (BEV) refers to a top-down viewing perspective, offering an overhead view of the surroundings.
% This perspective is advantageous as it provides minimal scale variation and occlusion, delivering a comprehensive view of the vehicle's environment, including the front, rear, sides, and top. 
% The BEV perspective naturally represents driving scenarios and is beneficial for subsequent downstream tasks.
Occupancy Prediction is a crucial task within the BEV perception framework, aiming to predict the occupancy status of the vehicle's surrounding environment. 
This includes forecasting the future positions and trajectories of dynamic objects such as pedestrians and other vehicles. 
Despite the growing trend of vision-based occupancy grid prediction under BEV perception, current research primarily focuses on outdoor highway driving scenarios, lacking studies on underground parking lot environments. 
Underground parking lots present unique challenges due to significant variations in lighting conditions, demanding robust perception systems capable of adapting to diverse illumination scenarios. 
Achieving superior perception performance in such environments remains a critical challenge for the autonomous driving industry.

The core objective of this study is to address the perception challenges faced by autonomous driving in adverse environments like basements. 
Initially, this paper commences with data collection in an underground garage. 
A simulated underground garage model is established within the CARLA simulation environment, and SemanticKITTI format occupancy ground truth data is collected in this simulated setting. Subsequently, the study integrates a Transformer-based Occupancy Network model to complete the occupancy grid prediction task within this scenario. 
A comprehensive BEV perception framework is designed to enhance the accuracy of neural network models in dimly lit, challenging autonomous driving environments. 
Finally, experiments validate the accuracy of the proposed solution's perception performance in basement scenarios.
The proposed solution is tested on our self-constructed underground garage dataset, SUSTech-COE-ParkingLot, yielding satisfactory results.

\end{abstract}

% Note that keywords are not normally used for peerreview papers.

% For peer review papers, you can put extra information on the cover
% page as needed:
% \ifCLASSOPTIONpeerreview
% \begin{center} \bfseries EDICS Category: 3-BBND \end{center}
% \fi
%
% For peerreview papers, this IEEEtran command inserts a page break and
% creates the second title. It will be ignored for other modes.
\IEEEpeerreviewmaketitle

%%%%%%%%%%%%%%%%%%%%%%%%%%%%%%%%%%%%%%%%%%%%%%
\section{Introduction}
%%%%%%%%%%%%%%%%%%%%%%%%%%%%%%%%%%%%%%%%%%%%%%

\subsection{Background}
The field of autonomous driving is undoubtedly one of the hottest topics in the current tech industry. 
Autonomous driving systems are complex operational systems that encompass all aspects required by AI, including perception, cognitive reasoning, decision-making, and control execution. 
The technical difficulty is extremely high, and because they involve the physical world, the margin for error is extremely low, aiming for even 100\% safety. 
Therefore, the requirements for technical capability boundaries are very stringent. 
Among the many tasks in autonomous driving, the perception and recognition task is the starting point for an intelligent agent to recognize the physical world, making it particularly important.

With the introduction of BEV (Bird's Eye View) into autonomous driving perception tasks, both academia and industry have been developed. 
BEV has seen such rapid development due to its distinct advantages: increasingly complex sensor calibration, data integration from multiple sensors, and the representation and processing of multi-source data need to be implemented within a unified BEV view.
BEV has advantages such as no occlusion and scale issues among objects, and objects and lane elements in the BEV view can be conveniently applied in subsequent modules, such as planning and control modules \cite{lan2023end}.

The main goal of BEV perception is to learn robust and generalized feature representations from camera and LiDAR inputs. 
A key issue is how to implement multi-source data fusion in the early or mid-stages of the solution. However, aligning and integrating multi-source data has always been a critical point with room for innovative development.
Compared to the LiDAR branch, which easily obtains 3D point cloud attributes, it is particularly challenging for the camera branch to obtain 3D spatial information using monocular or multi-view settings.
From this perspective, there is significant room for development in obtaining spatial information from visual solutions.

Currently, vision-based BEV solutions still lag significantly behind BEV LiDAR solutions.
The performance gap between vision algorithms and LiDAR algorithms on the nuScenes dataset exceeds 20\%, and on the Waymo dataset, this gap reaches 30\%. 
From an academic perspective, the key to vision algorithms achieving performance on par with LiDAR is better understanding the view transformation from 2D external input to 3D geometric input. 
From an industrial perspective, equipping an autonomous vehicle with a set of LiDAR devices is far more expensive than vision software. 
This demand drives the development of vision-based autonomous driving algorithms. 
Among various perception tasks based on vision and LiDAR, vision perception has the potential to compete with LiDAR perception. 
Moreover, current research and development indicate that BEV feature representation is also suitable for multi-camera input. 
These factors collectively promote the development of BEV visual perception.

In the field of vision-based BEV perception, lighting issues are particularly challenging. 
The lighting quality in different scenarios affects the images captured by cameras differently, significantly impacting the effectiveness of BEV perception.
Basement scenarios are important for autonomous driving tasks, characterized by poor lighting conditions \cite{lan2024sustechgan}.
Despite the ongoing development of BEV perception technology, current open datasets mainly feature outdoor scenarios, and existing solutions still struggle to adapt well to basement environments. 
Experiments in this study found that state-of-the-art (SOTA) algorithm models from major public datasets perform poorly in basement scenarios. 
Therefore, this paper aims to design a solution to optimize BEV perception performance in basement scenarios.

\subsection{Main Work}
\textbf{Construction of Simulation Environment:} 
Utilizing the Unreal Engine-based simulation platform CARLA, we simulate a realistic scenario of the SUSTech-COE underground parking lot, constructed using 3D modeling software. 
By meticulously studying the layout and features of the underground garage based on engineering CAD drawings, we aim to provide a highly realistic simulation environment to ensure the quality of subsequent data collection and driving scene experiments.

\textbf{Data Collection and Scripting:} 
We develop corresponding scripts within the constructed simulation environment using the Python API interface provided by the CARLA simulator to facilitate data collection.
Specifically, we configure vehicle and sensor parameters within the CARLA simulation world, control vehicle and sensor behaviors through scripts, and collect a large amount of camera image information, sparse point cloud data, and semantic information. 
This data collection process provides the foundation for capturing dynamic changes in driving scenes within the virtual environment.

\textbf{Multi-frame Fusion for Point Cloud Densification:} 
Using the collected data, we apply a multi-frame fusion algorithm to densify sparse point cloud data and its corresponding semantic information \cite{lan2022semantic,liu2022towards}. 
By integrating multiple frames of data into a single frame's point cloud, we significantly improve the density and accuracy of the point cloud. 
This step aims to generate high-quality point cloud data for training subsequent algorithm models.

\textbf{Occupancy Grid Prediction with Vision-based Semantic Scene Completion Model:} 
Finally, we employ a Transformer-based vision semantic scene completion model to predict occupancy grids. 
This model, leveraging the powerful feature extraction capabilities of Transformers and featuring a novel two-stage design, achieves high-performance perception while reducing the computational resources required for training.

Through these steps, we have established a complete workflow from simulation environment construction, data collection, and processing to final occupancy grid prediction, validating the effectiveness and practicality of the proposed method.

%%%%%%%%%%%%%%%%%%%%%%%%%%%%%%%%%%%%%%%%%%%%%%
\section{Related Work}
\label{sec:related_work}
%%%%%%%%%%%%%%%%%%%%%%%%%%%%%%%%%%%%%%%%%%%%%%

\subsection{Vision-based 3D Perception}
Vision-based solutions have garnered widespread attention in the field of autonomous driving due to their low cost, ease of deployment, and wide applicability. 
Additionally, cameras can provide rich visual attributes of the scene to aid in overall scene understanding for vehicles.
Recently, there has been a surge of work on 3D object detection or segmentation from RGB images. 
Inspired by DETR's \cite{carion2020end} success in 2D detection, DETR3D \cite{wang2022detr3d} links learnable 3D object queries with 2D images through camera projection matrices, achieving end-to-end 3D bounding box prediction without the need for non-maximum suppression (NMS).
M$^2$BEV \cite{xie2022m} has also investigated the feasibility of simultaneously running multi-task perception based on BEV features. 
BEVFormer \cite{li2022bevformer} proposes a spatiotemporal fusion Transformer structure that aggregates BEV features from both current and previous frames using deformable attention \cite{zhu2020deformable}. 
Compared to object detection tasks, occupancy grid prediction tasks can represent the occupancy status of each small voxel unit instead of assigning fixed-size bounding boxes to objects. 
This helps autonomous driving perception tasks better identify objects with irregular states. 
Compared to 2D BEV schemes, 3D voxelized scene representations contain more information, which is more accurate in complex driving environments (such as rugged roads). 
sense voxel semantics can provide a more comprehensive 3D scene representation, and efficiently and accurately achieving this goal through vision-based approaches is challenging.

\subsection{Semantic Scene Completion}
3D semantic scene completion presents a significant challenge for autonomous driving vehicles due to the limited range of sensors.
SSCNet \cite{song2017semantic} first defined the task of semantic scene completion, which involves jointly inferring the completion of the scene based on geometric and semantic information given incomplete visual observations. 
In recent years, with the release of the SemanticKITTI dataset \cite{behley2019semantickitti}, the task of semantic scene completion in large-scale outdoor scenes has gained attention.

Completing semantic scene information based on sparse observations is ideal for autonomous driving vehicles. 
Vehicles with this capability can obtain dense 3D voxelized semantic representations of scenes, which can aid in reconstructing 3D static maps and perceiving dynamic objects. 
However, large-scale 3D semantic scene completion in driving scenarios is still in its early stages of exploration and development. 
Existing work typically relies on point cloud data as input \cite{rist2021semantic}. 
In contrast, the recent MonoScene \cite{cao2022monoscene} study explored semantic scene completion based on monocular images. 
This work proposed feature projection from 2D to 3D and utilized continuous 2D and 3D Unet to achieve 3D semantic scene completion. 
However, the feature projection from 2D to 3D is susceptible to false features introduced by unoccupied positions, and multi-layered 3D convolution can reduce system efficiency.

\subsection{Occupancy Prediction}
3D occupancy prediction, as a hot topic in perception tasks, has been widely discussed recently. 
It involves predicting environmental occupancy and semantic information at the finest granularity, exhibiting good scalability and adaptability to downstream tasks. 
In recent works, solutions attempt to predict semantic occupancy solely from image inputs, while OpenOccupancy \cite{wang2023openoccupancy} employs a multisensor fusion approach to tackle this task.
TPVFormer \cite{huang2023tri} combines inputs from surrounding multi-camera views and elevates features to a three-view space using a Transformer-based \cite{vaswani2017attention} approach. 
Due to its training with sparse LiDAR point clouds for supervision, its predictions are also sparse. 
Similarly, SurroundOcc \cite{wei2023surroundocc} adopts a Transformer-based method to generate 3D voxel features at multiple scales and combines these features through deconvolution upsampling. 
Additionally, this work proposes a method to extract dense semantic occupancy information from sparse LiDAR data for supervision, enabling dense prediction results. 
However, the inference process of these methods is time-consuming \cite{lan2018real,lan2019evolving} and far from meeting the real-time requirements of autonomous driving perception. 
For example, the network in SurroundOcc requires over 300 milliseconds for a single inference, which is not ideal for real-time applications \cite{yi2024key}. 
Although most methods directly transform images into dense 3D voxel representations using carefully designed techniques to achieve better results, the trade-off in inference speed may not necessarily yield the optimal solution.

%%%%%%%%%%%%%%%%%%%%%%%%%%%%%%%%%%%%%%%
\section{Methodology}
\label{sec:methodology}
%%%%%%%%%%%%%%%%%%%%%%%%%%%%%%%%%%%%%%%

Due to the lack of underground parking scenarios in publicly available datasets for autonomous driving, this study aims to explore the performance of BEV perception occupancy grid algorithms in such environments. 
To achieve this, we first use the modeling software Maya to construct a corresponding 3D map based on the CAD drawings of the SUSTech underground parking lot, thus virtualizing the real-world scenario. 
Next, we utilize the CARLA simulator to generate entities such as the ego vehicle and sensors within this map, setting vehicle driving paths and sensor parameters to obtain corresponding image and point cloud information from the simulation map, approximating real-world scene data.

Subsequently, we preprocess the initially collected data, densifying the sparse point cloud through multi-frame fusion and voxelizing it to obtain ground truth data suitable for model training. 
Finally, we use the collected data as input and the preprocessed data for supervision to train the occupancy grid prediction network model, VoxFormer \cite{li2023voxformer}, and test its perception performance in the underground parking lot.

By following this methodology, we aim to evaluate the effectiveness of the VoxFormer model in confined and dimly lit environments, ensuring the generated dataset and simulation accurately reflect the challenges present in real-world underground parking scenarios.

\subsection{Simulation Scene Modeling}
This paper uses the underground parking lot of the SUSTech School of Engineering as the real-world basis to construct a CARLA map based on Unreal Engine, named SUSTech COE ParkingLot. 
% The CAD drawing of this parking lot is shown in \autoref{fig:parkinglot}.

The aim of this study is to enhance the perception performance of occupancy grid network models in confined and dimly lit spaces such as underground parking lots. 
To achieve this, we first need to generate targeted data.

Data set construction can be approached in two ways: one is through manual collection followed by semantic annotation, and the other is by creating twin scenes using a simulator, where each scene element is pre-classified and true-value data is generated through a 3D virtual engine \cite{lan2023virtual}. 
While manual collection and subsequent annotation can yield more realistic data, it also involves a significant amount of work. 
Given the project's requirement for extensive semantic annotations, with each image taking approximately three minutes to annotate, we constructed the dataset through simulator-based twin scenes.

The specific construction process is as follows: First, based on the construction drawings of the target underground parking lot, we meticulously mark the position and size of each unit element. 
Then, using industrial 3D modeling software such as Maya and 3Ds Max, we transform the 2D construction CAD drawings into a 3D model by establishing the z-axis direction in the 3D space, creating a 3D scene. 
Finally, we use the Unreal Engine to apply texture mapping and semantic information to the 3D scene, completing the construction of the entire simulation twin scene.

The constructed simulation scene will be used for further data collection and simulation road testing tasks.

\subsection{Data Collection}
The collected data is saved in the SemanticKITTI format, and images captured by stereo cameras for each frame are also stored. 
This collected data then undergoes ground truth generation, where point cloud information is voxelized and semantic scene completion is performed using appropriate algorithms.

\subsubsection{Scenario Settings}
In the CARLA environment, we configure the predefined driving scenarios to replicate real-world scene information accurately.

\textbf{Map Settings:}
In CARLA, we create a map for the SUSTech COE ParkingLot (SUSTech College of Engineering underground parking lot) to replicate the real-world environment. 
Since underground parking lots are generally unaffected by weather conditions and have dim lighting, we set the world weather to cloudy and overcast.

\textbf{Frame Rate Settings:}
We set the simulator's frame rate to 1000 frames per second and configured the communication mode between the server and client to synchronous mode. 
In synchronous mode, the flow of time in the CARLA world is controlled by external scripts, meaning that the client running the Python code takes control and instructs the server when to update the next frame.

\subsubsection{Ego Vehicle Settings}
We configure the basic information for the vehicle and make different positional adjustments to meet data collection requirements.

\textbf{Spawn Point Settings:}
In the CARLA world, we create an autonomous vehicle (Ego Vehicle) using the Tesla Model 3 series and select different spawn points based on the data collection requirements. 
The driving scenario in an underground parking lot is more complex compared to open roads, with more lanes and obstacles such as walls and pillars. 
To collect complete information about the entire parking lot, the vehicle needs to navigate through every lane, covering all corners and areas obscured by walls and pillars to ensure the quality of data collection. 
Therefore, it is essential to carefully choose the spawn points of the ego vehicle within the underground parking lot. 
For this purpose, we manually select 22 different spawn points to cover all road segments within the parking lot.

\textbf{Anti-collision Settings:}
During the vehicle spawning process, CARLA suspends the vehicle in the air to prevent collisions with objects along its z-axis, such as the ground, which could lead to issues with the physics engine. 
While waiting for the vehicle to land, data collection cannot begin. 
Therefore, it is necessary to calculate the free fall time of the vehicle and wait for the corresponding number of frames before data collection can commence.

\subsubsection{Sensor Settings}
On the autonomous vehicle, we attach relevant sensors such as cameras, LiDAR, and set their corresponding parameters \cite{xu2019online}. 
We reference the sensor parameter settings from the KITTI official specifications and make numerical adjustments based on the situation in CARLA. 
In the real world, these parameters need to be determined through camera calibration. 
We can directly set their corresponding values in an ideal simulated environment \cite{lan2019simulated}.

\textbf{Camera Settings:}
We employ PointGray Flea2 color and grayscale cameras, with parameters set as shown in \autoref{tab:camera_intern}.
    
\begin{table}[!ht]  \centering \small
\setlength\tabcolsep{1pt} \renewcommand{\arraystretch}{1.0}
    \caption{Internal Parameter of Camera}
    \label{tab:camera_intern}  
    \begin{tabular}{cccccc}
        \hline\noalign{\smallskip}	
        \textbf{Camera} & \textbf{Resolution} & \textbf{FOV} & \textbf{Refresh} & \textbf{Gamma} & \textbf{Shutter}\\
        \noalign{\smallskip}\hline\noalign{\smallskip}
        Left RGB & 1240*370 & 80 & 10 & 2.2 & 1000\\
        Right RGB & 1240*370 & 80 & 10 & 2.2 & 1000\\
        Left Grayscale & 1240*370 & 80 & 10 & - & 1000\\
        Right Grayscale & 1240*370 & 80 & 10 & - & 1000\\
        \noalign{\smallskip}\hline
    \end{tabular}
\end{table}

\textbf{Lidar Settings:}
We use the Velodyne-HDL-64E LiDAR with parameters set as shown in \autoref{tab:lidar_intern}.
    
\begin{table}[!ht]  \centering  \small
    \caption{Internal Parameters of Lidar}
    \label{tab:lidar_intern}
    \begin{tabular}{cc}
        \toprule
        \textbf{Parameter} & \textbf{Value} \\
        \midrule
        Number of Channels & 64 \\
        Range & 100 \\
        Point Cloud Frequency & 2200000 \\
        Rotation Frequency & 10 \\
        Upper Angle & 2 \\
        Lower Angle & -24.8 \\
        \bottomrule
    \end{tabular}
\end{table}

\textbf{Sensor Parameter Settings:}
The external parameter settings of all sensors relative to the ego vehicle coordinate system are shown in \autoref{tab:sensor_extern}.

\begin{table}[!ht] 
    \centering 
    \small
    \caption{External Parameters of Sensors, which Position is performed in (x, y, z), and Rotation is performed in (pitch, yaw, roll)}
    \label{tab:sensor_extern}  
    \begin{tabular}{ccc} 
        \toprule
        \textbf{Sensor} & \textbf{Position} & \textbf{Rotation} \\
        \midrule
        Left RGB Camera & (0.30, 0, 1.70) & (0, 0, 0) \\
        Right RGB Camera & (0.30, 0.50, 1.70) & (0, 0, 0) \\
        Left Grayscale Camera & (0.30, 0, 1.70) & (0, 0, 0) \\
        Right Grayscale Camera & (0.30, 0.50, 1.70) & (0, 0, 0) \\
        Top Lidar & (0, 0, 1.80) & (0, 180, 0) \\
        \bottomrule
    \end{tabular}
\end{table}

\subsubsection{Data Acquisition and Storage}
Once the vehicle and sensor parameters are properly set, the core part of data collection can begin:

Construct the corresponding file directory structure based on the SemanticKITTI dataset structure, with the addition of a calib.txt file, representing the parameter information of the vehicle-mounted sensors and the affine transformation relationship, on the basis of the file structure of SemanticKITTI.

Save the data obtained from the built-in LiDAR sensors in CARLA in the .bin file format under the velodyne folder. 
The required point cloud data is saved as the following vectors. 
Since the semantic annotations for each entity in the world in CARLA are not the same as SemanticKITTI, semantic remapping is also needed. 
The indices corresponding to different entity categories in CARLA are replaced with those in KITTI, and saved in the label file format under the labels folder.

In addition, for each sensor's parameters, it is necessary to perform affine transformations to different coordinate systems and calculate intrinsic, extrinsic, and projection matrices. 
Finally, these parameters are output to the camera calibration parameter file ‘calib’ and the pose parameter file ‘poses’ to prepare the data for subsequent ground truth generation tasks.

\begin{enumerate}
    \item In the ‘calib’ file, the first four lines contain the camera calibration data, each consisting of a rectified 3x4 projection matrix. Specifically:
    \begin{itemize}
        \item Lines 1 and 3 represent the matrices corresponding to the left-side RGB and grayscale cameras, respectively.
        \item Lines 2 and 4 represent the matrices for the corresponding right-side RGB and grayscale cameras.
        \item These matrices can map a point from the camera coordinate system to the image coordinate system of the corresponding camera.
        \item The last line consists of the transformation matrix that rotates the coordinate system from the left-side RGB camera to the top-mounted LiDAR coordinate system. This matrix maps a point from the LiDAR coordinate system to the left-side RGB camera coordinate system.
    \end{itemize}
    \item The ‘poses’ file records the pose information of the camera throughout the data collection process. 
    Each line is a 3x4 transformation matrix representing the pose of the current frame relative to the starting frame in the left camera coordinate system. 
    This matrix maps a point from the current frame to the starting frame's coordinate system.
\end{enumerate}

\begin{figure}[!ht]  \centering
    \includegraphics[width=0.48\textwidth,trim={15 0 15 0},clip]{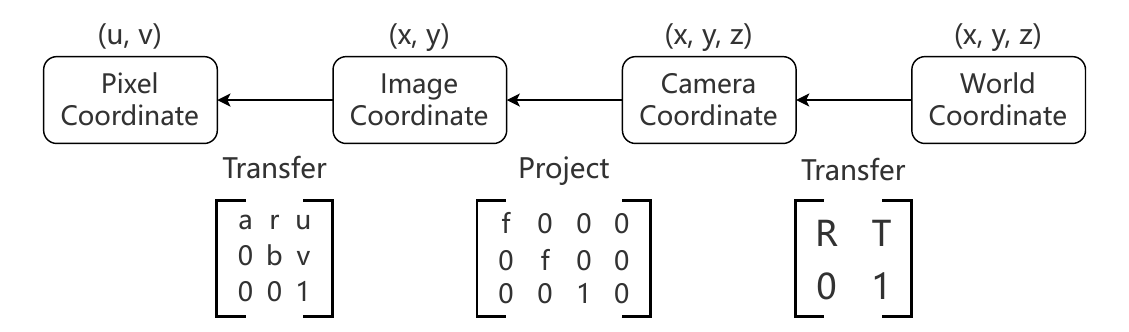}
    \caption{The coordinates to be transformed are augmented with an additional dimension, and the rotation matrix is zero-padded to form a four-dimensional homogeneous matrix.
    By left-multiplying the corresponding matrices from right to left, one can project the coordinates on the right to the corresponding coordinate system on the left. 
    In this context, the rigid transformation matrix $\mathbf{R} $ represents the rotation matrix projecting a point from the world coordinate system to the camera coordinate system. 
    This matrix is also the inverse of the pose matrix of the camera relative to the world coordinate system. 
    The translation vector $\mathbf{T} $ indicates the corresponding translation vector, representing the distance from the origin of the world coordinate system to the origin of the camera coordinate system.}
    \label{fig:transfer}
\end{figure}

The remaining transformation relationships can be deduced similarly. 
The world coordinate system describes the positions of objects in the real world and is typically used as the reference frame for the entire scene. 
The camera coordinate system is an internal coordinate system of the camera used to describe the image information within the camera sensor. 
Its origin is at the camera's optical center, with the z-axis coinciding with the optical axis (pointing directly in front of the camera) and the focal length $f $ being the distance between the optical center and the image coordinate system. 

The image coordinate system is a two-dimensional coordinate system on the camera's imaging plane, representing pixel positions on the camera's captured image. 
Its origin is at the intersection of the optical axis and the imaging plane, with the x and y axes parallel to the horizontal and vertical directions of the camera coordinate system, respectively. 

The pixel coordinate system represents the image as displayed on electronic devices, with its origin at the top-left corner of the image.
The x and y axes are parallel to the corresponding axes of the image coordinate system.

\begin{figure}[!ht]  \centering
    \includegraphics[width=0.48\textwidth]{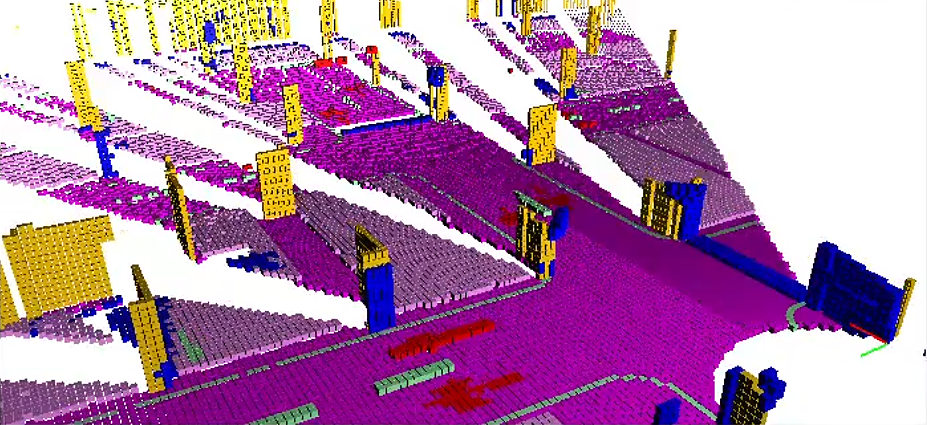}
    \caption{Incomplete Data Visualization, Which yellow voxels refer to wall, purple voxels refer to road, blue voxels refer to other stable features,  green and red voxels refer to traffic line on the road.}
    \label{fig:uncompleted}
\end{figure}

After processing the aforementioned steps, we can collect the initial data, which is incomplete data, as shown in \autoref{fig:uncompleted} At this stage, the data does not yet meet the quality requirements and must undergo further preprocessing to generate ground truth before it can be used for model training.

The ground truth generated during the data preprocessing stage is primarily used to train the occupancy grid prediction model.
This process is divided into two stages, each requiring different types of ground truth data. 
The ground truth's specific content depends on the model's requirements at each stage.

\subsection{Data Preprocessing}
In this phase, the preliminary data collected in the previous steps needs to undergo a series of preprocessing steps to provide the input data required for the occupancy grid prediction model inference and the ground truth needed for training. 
Since the ground truth generation for the first phase depends on the ground truth from the second phase, this paper will first describe the method for generating the ground truth for the second phase.

\subsubsection{Second-stage Truth Generation}
\begin{figure}[!ht]  \centering
    \includegraphics[width=0.485\textwidth]{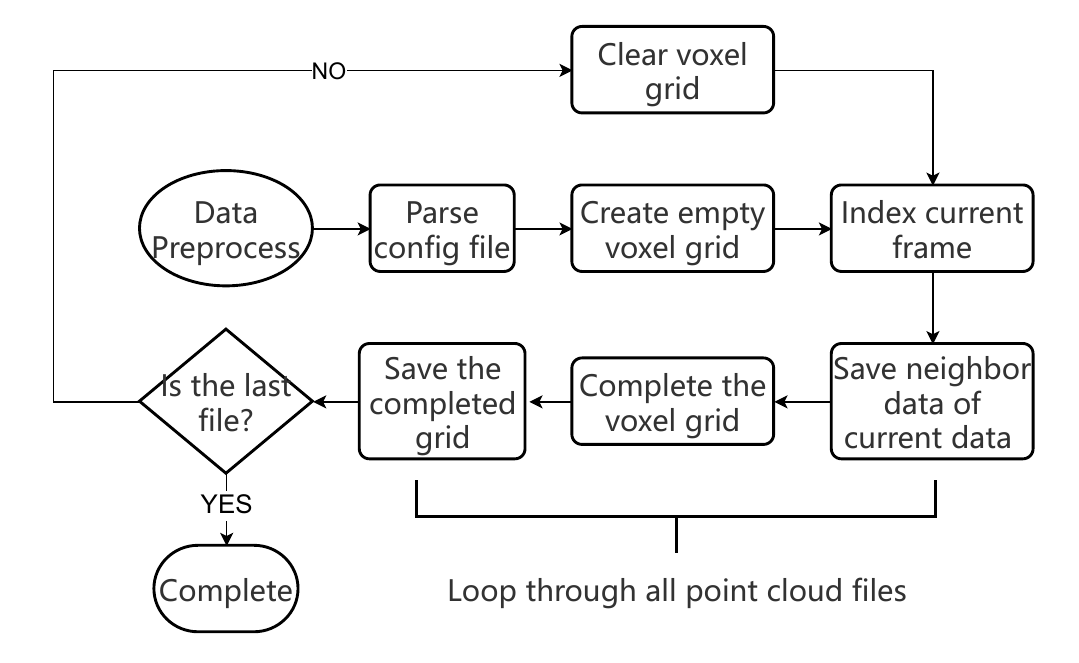}
    \caption{Flowchart of the completion algorithm}
    \label{fig:voxel-complete}
\end{figure}

Based on the trajectory information depicted in the poses file, the sparse point clouds in the point cloud data are completed through multi-frame fusion and voxelized.
These are then mapped onto a voxel grid map, which is a fixed-size cubic structure composed of multiple voxels, and saved. 
The generation of this ground truth can supervise the training of the occupancy grid prediction network model. 
The specific process is illustrated in \autoref{fig:voxel-complete}.

For the voxel completion algorithm in the second stage, the specific steps are as follows:
First, the configuration in the config file is parsed, focusing on parameters such as priorScan (number of preceding scans to consider) and pastScan (number of subsequent scans to consider). 
These parameters control the number of preceding and subsequent frames the algorithm must consider during the voxel completion process. 
Modifying these parameters in the config file allows the algorithm to produce results that better meet the requirements.

Next, the VoxelGrid class objects \cite{lan2022class,gao2021neat}, priorGrid and pastGrid, are constructed to save the completed voxel maps. 
They are initialized as empty 256×256×32 three-dimensional arrays.

Finally, all point cloud files are iterated over.
For each point cloud file $F_t $, the point clouds and labels from preceding and subsequent frames, as specified in the config settings, are fused into the same VoxelGrid.

In this step, all point cloud and label data within the range $[t-priorScan, t]$ are first read and saved into the variables prior points and prior label.

Then, the point cloud and label data for the past frames following time $t $ are saved in the same manner.

\begin{figure}[!ht]  \centering
    \includegraphics[width=0.5\textwidth]{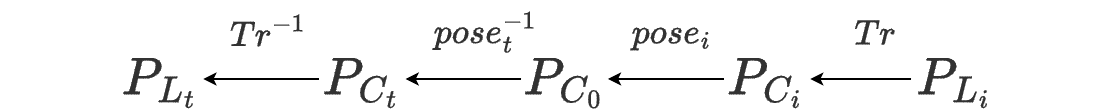}
    \caption{where $P_C $ represents a point in the camera coordinate system, with the subscript indicating the corresponding frame number.
    The transformation matrices $Tr $ and $Tr^{-1} $ represent the matrices for projecting a point from the LiDAR coordinate system to the left-front camera coordinate system and their inverses, respectively. 
    $pose_i $ and $pose_t^{-1} $ represent the matrices for projecting a point from the left-front camera coordinate system at frame $i $ to the coordinate system at frame 0, and for projecting a point from the left-front camera coordinate system at frame 0 to the coordinate system at frame $t $, respectively. 
}
    \label{fig:transfer-coord}
\end{figure}

The data read in the previous two steps are then fused into the corresponding VoxelGrid. 
First, all point clouds need to be projected onto the same frame. 
Next, each voxel in the grid is assigned its corresponding semantic label:

For the point clouds used for filling, affine transformations between coordinate systems are required using the calib and poses files from the data collection phase: Let frame $i $ be the frame of the point cloud sequence used to fill the voxel grid at frame $t $. 
For a point in the point cloud of frame $i$, it needs to be transformed from the coordinate system of the LiDAR at frame $i $ to that of the LiDAR at frame $t$. 
Let its coordinates be $P_{L_i} $ and its projected coordinates be $P_{L_t} $. 
The transformation process is shown in \autoref{fig:transfer-coord}.

Through this affine transformation, we can calculate the representation of a point in the LiDAR coordinate system at frame $i $ as a point in the LiDAR coordinate system at frame $t $. 
Consequently, all the point clouds and semantic labels from the frames involved in the completion process can be fused into the frame at $t $.

This paper adopts a method of counting the semantics of the point clouds included within a voxel unit to assign semantic labels. 
The semantic with the highest proportion is directly assigned as the semantic of the current voxel. 

\begin{figure}[!ht]  \centering
    \includegraphics[width=0.48\textwidth]{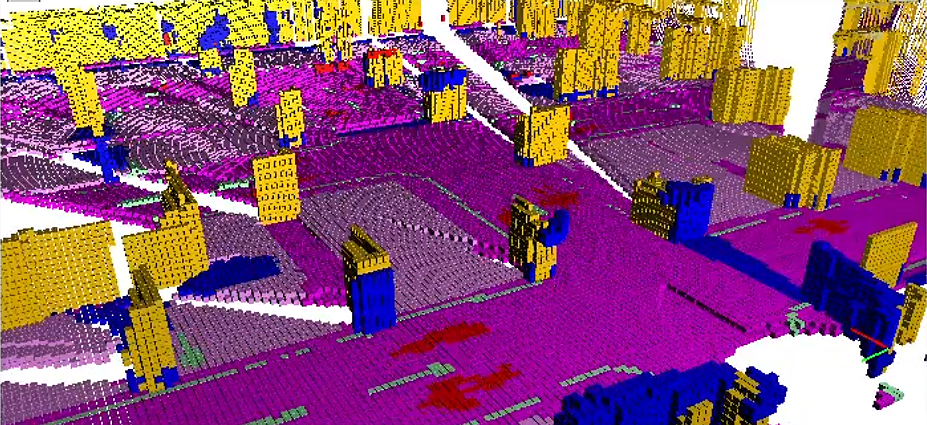}
    \caption{Completed Data Visualization, which colored voxels refer to the same semantics in \autoref{fig:uncompleted}.}
    \label{fig:completed}
\end{figure}

After processing the above steps, we can obtain the completed voxel grid, as shown in \autoref{fig:completed}. 
This grid can be used as ground truth for training in the second stage of occupancy grid prediction.

After completing the sparse point cloud and moving it to a dense point cloud, the corresponding voxelization algorithm step can be performed.
Specifically, for each voxel region, we count the semantic labels of all the point clouds within its range, and then assign the most frequent semantic label to the current voxel unit.
This results in the voxelization of the corresponding point cloud.

\subsubsection{First-stage Truth Generation}
\begin{figure}[!ht]  \centering
    \includegraphics[width=0.48\textwidth]{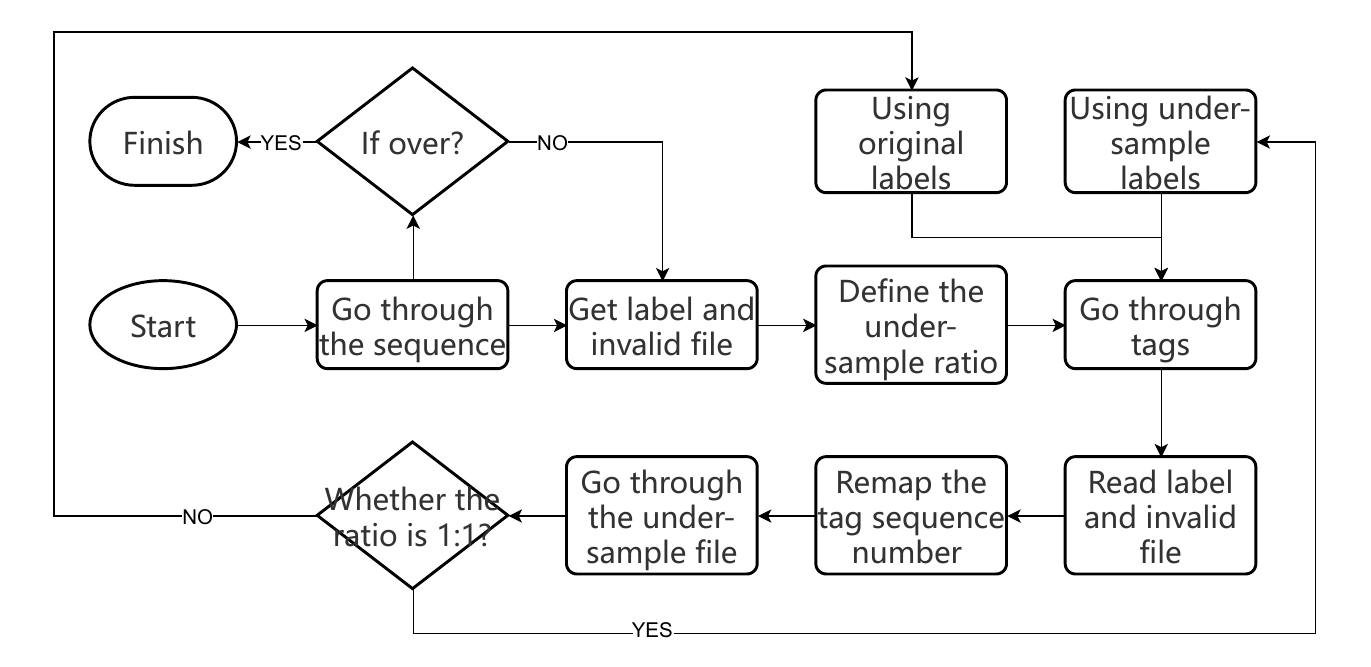}
    \caption{Flowchart of the Under-sample algorithm}
    \label{fig:under-sample}
\end{figure}

The first stage ground truth is derived from the second stage ground truth with some modifications. 
First, the discrete semantic indices in the first stage voxel are mapped to a continuous numerical range. 
Subsequently, the grid with a resolution of 256×256×32 is downsampled. 
The downsampling method involves counting the number of semantic values equal to 0 or 255 within multiple high-resolution voxel units corresponding to each low-resolution voxel unit. 
The voxel is marked as occupied if this count is below a certain threshold. 
This process yields a class-agnostic occupancy grid with a resolution of 128×128×16. The basic flowchart is shown in \autoref{fig:under-sample}.

%%%%%%%%%%%%%%%%%%%%%%%%%%%%%%%%%%%%%%%%%%%%%%
\section{Experiments}
\label{sec:experiments}
%%%%%%%%%%%%%%%%%%%%%%%%%%%%%%%%%%%%%%%%%%%%%%

To validate the effectiveness of the model, we conducted comparative experiments. 
First, we used a model trained solely on the SemanticKITTI dataset to predict outcomes in the CARLA simulator's underground parking lot, obtaining the initial prediction results.
Then, we fine-tuned this model using the data we collected independently, yielding another set of prediction results. 
By comparing the evaluation metrics of both predictions and combining this with visual inspection, we performed a preliminary assessment of the model's perception capabilities.

\subsection{Environment Configuration}
Since the inference and training of the occupancy grid prediction model require GPU computational support, we used SSH to connect to a remote server cluster for model inference and training. 
The specific hardware and software environment are as follows:
\begin{itemize}
    \item \textbf{Programming Languages:} Python 3.8, C++
    \item \textbf{Deep Learning Framework:} PyTorch 1.9.1
    \item \textbf{Operating System:} Ubuntu 22.04
    \item \textbf{Graphics Cards:} 2 NVIDIA L40S GPUs
    \item \textbf{Development Environment:} PyCharm Professional, Docker
    \item \textbf{Python Packages:}
    \begin{itemize}
        \item torchvision 0.10.1
        \item CUDA Toolkit 11.1
        \item torchaudio 0.9.1
        \item mmcv 1.4.0
        \item mmdet 2.14.0
        \item mmsegmentation 0.14.1
        \item timm
    \end{itemize}
    \item \textbf{Compilation Environment:} gcc 6.2, NVCC 11.1
\end{itemize}

We chose Miniconda for Python package management. 
Miniconda is a lightweight distribution of Anaconda, serving as a package and environment manager for data science and machine learning \cite{lan2022time,lan2021learning,lan2021learning2}. 
Unlike Anaconda, which includes numerous pre-installed packages, Miniconda only contains the essential components, making installation and updates faster and more efficient.

Additionally, all experimental environments related to the model were deployed in Docker containers on the server cluster. 
Docker is a containerization platform that allows developers to package all environment dependencies and applications into a portable container. 
This container can run on any operating system, preventing issues related to OS environment conflicts and ensuring the program can run on different platforms.

Here’s a brief on why using Miniconda and Docker:
\begin{itemize}
    \item \textbf{Miniconda:} Provides a minimal environment to install only necessary packages, leading to efficient resource use.
    \item \textbf{Docker:} Ensures that the entire experimental setup, including dependencies and configurations, is encapsulated in a container, enhancing reproducibility and portability across different environments.
\end{itemize}

\subsubsection{CARLA Simulator Configuration}

\begin{figure}[!ht]  
\centering
    \includegraphics[width=0.48\textwidth]{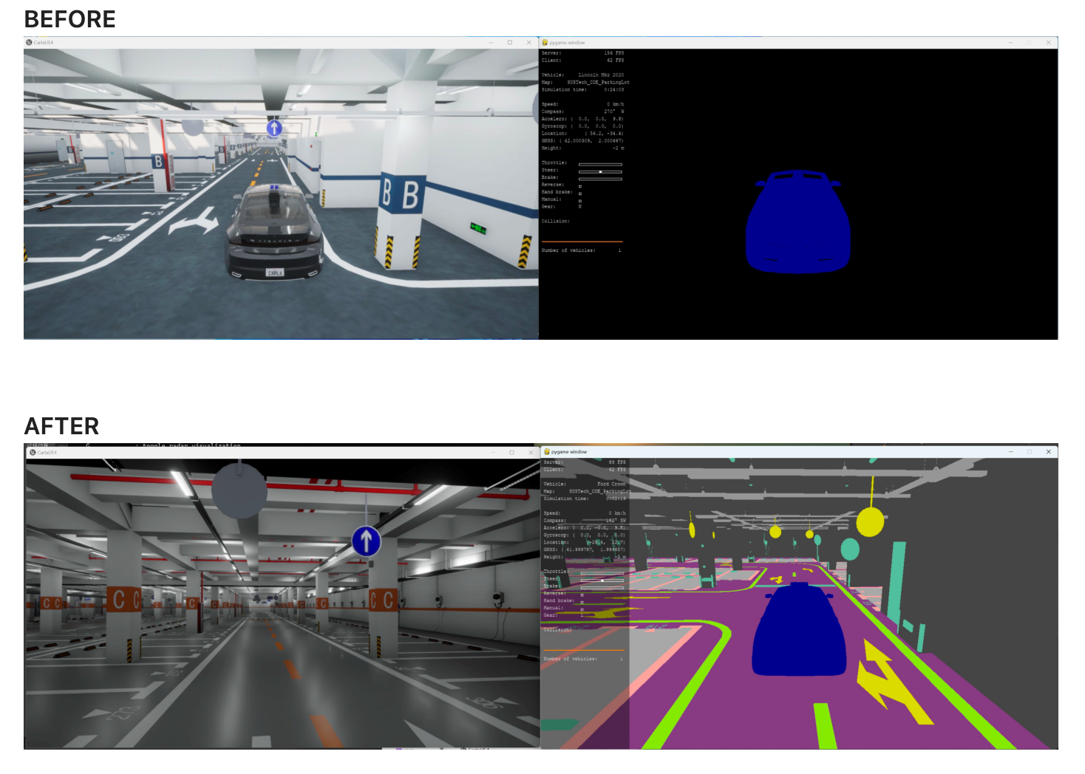}
    \caption{Comparison of Semantic Addition Effects: \textbf{Before} shows what the scene looks like when semantics have not been added to the map, and the light and shadow details of the world are poor. \textbf{After} shows the scene after the semantics have been added, different colors in the picture represent different meanings, and the light and texture of the world are good.}
    \label{fig:change}
\end{figure}

Since the experimental scenario conceived in this paper is an underground parking garage, the official CARLA simulator does not include certain semantics specific to such environments, such as parking spaces and parking lines. 
Therefore, we extended the official version by adding these semantics and compiled them into the CARLA simulator. 
The specific implementation effect is shown in \autoref{fig:change}.

As can be seen, before adding the semantics, the information feedback in CARLA only included the semantics of the ego vehicle, with no annotations for other elements. 
After compiling the semantics into the simulator, it displays rich semantic information, where each color corresponds to a different semantic category.

\subsubsection{Dataset}
The RGB camera data collected in the underground parking garage dataset is aligned with the stereo camera format in the KITTI Odometry dataset \cite{geiger2012we}. 
The collected images are shown as \autoref{fig:carla_rgb}:
\begin{figure}[!ht]
    \centering
    \includegraphics[width=0.48\textwidth]{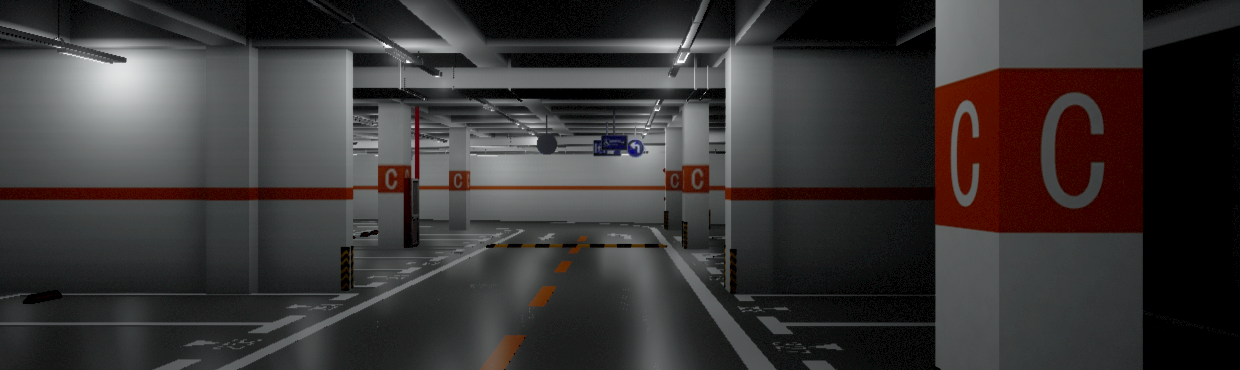}
    \caption{RGB camera data collection result}
    \label{fig:carla_rgb}
\end{figure}

\begin{figure}[!ht]   \centering
    \includegraphics[width=0.485\textwidth]{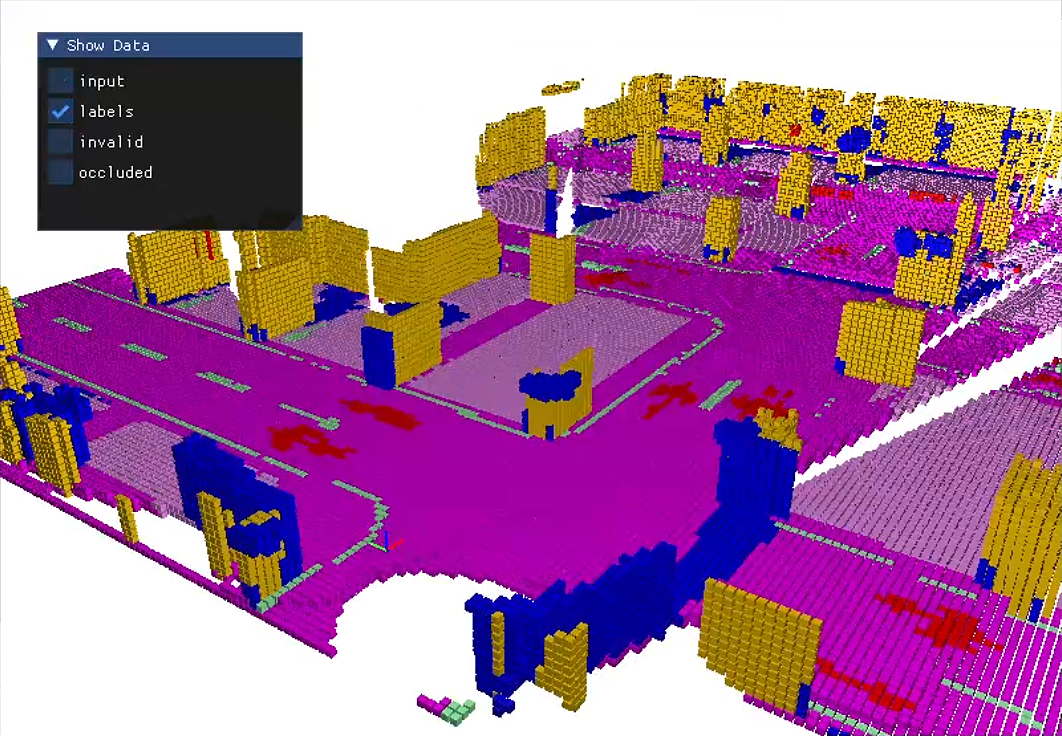}
    \caption{Visualization of label file result: Grid map after semantic completion, color semantics correspond to \autoref{fig:uncompleted}.}
    \label{fig:label_visualization}
\end{figure}

In addition, other data that needs alignment includes the ground truth voxel used for the second stage of training. 
The visualized files after alignment are as \autoref{fig:bin_visualization} and \autoref{fig:label_visualization}:
\begin{figure}[!ht]  \centering
    \includegraphics[width=0.485\textwidth,trim={10 10 10 10},clip]{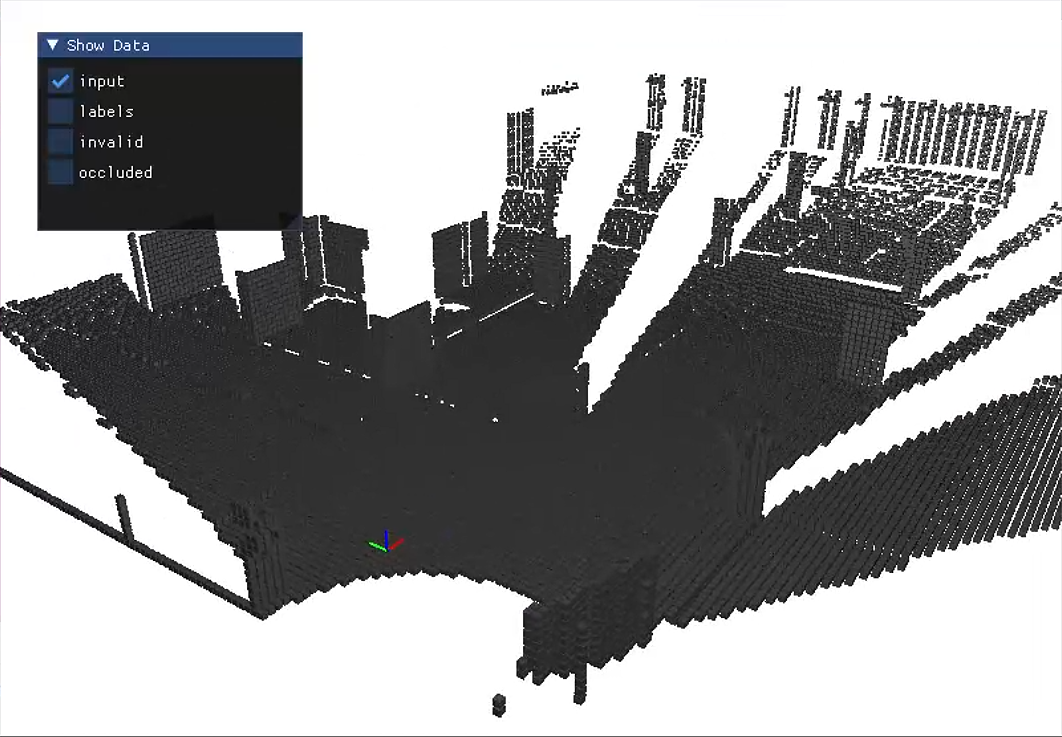}
    \caption{Visualization of .bin file result: The black voxels represent occupied grids.}
    \label{fig:bin_visualization}
\end{figure}

\subsubsection{Implement Details}
In the first stage, VoxFormer utilizes MobileStereoNet for direct depth estimation, enabling low-cost generation of pseudo-LiDAR point clouds. 
The occupancy prediction network used for depth correction is adapted from LMSCNet \cite{roldao2020lmscnet}, a lightweight 2D convolutional neural network \cite{lan2022vision,lan2024dir}.
During training, the input is a voxelized pseudo point cloud of size $256 \times 256 \times 32$, and the output is an occupancy grid of size $128 \times 128 \times 16$.

In the second stage, VoxFormer employs ResNet-18 \cite{he2016deep} to extract features from 2D images. 
These features are then fed into an FPN \cite{lin2017feature} to generate a feature map of size $\frac{1}{16}$ of the input image. 
The feature dimension is set to $d=128$; the number of deformable cross-attention layers is 3, and the number of deformable self-attention layers is 2. 
In this work, 8 sampling points are selected around the reference point. 
The final fully connected layer projects the 128-dimensional features into 20 categories.

For both the first and second stages of training, 24 epochs were used, with a learning rate set to $2 \times 10^{-4}$.

\subsubsection{Evaluation}
In this paper, the evaluation of semantic scene completion \cite{song2017semantic} adopts Intersection-over-Union (IoU), irrespective of the assigned semantic labels. 
The performance of semantic segmentation is evaluated using mean Intersection-over-Union (mIoU). 
The performance of the semantic scene completion algorithm is typically measured by the IoU between the ground truth and the prediction results.

Intersection-over-Union (IoU) is the ratio of the intersection to the union of the predicted and actual samples, representing the number of correctly predicted points for a given label divided by the sum of all points predicted as that label and the actual points for that label. 
For the scene completion part in this paper, the occupancy grid is composed of 0-1, so in practice, the calculation of IoU is evaluated only on a binary occupancy grid. 
This metric reflects the accuracy of the semantic segmentation results for each category, which can be calculated by:
\begin{equation} \centering 
\text{IoU} = \frac{X \cap Y}{X \cup Y} = \frac{\text{TP}}{\text{FP} + \text{TP} + \text{FN}}
\label{eq:iou}
\end{equation}

For the task of semantic scene completion, after obtaining the IoU, the overall performance is evaluated by calculating the mean Intersection-over-Union (mIoU) for all categories as:
\begin{equation} \centering 
\text{mIoU} = \frac{1}{k+1} \sum_{i=0}^{k} \frac{\text{TP}}{\text{FP} + \text{TP} + \text{FN}}
\label{eq:miou}
\end{equation}

Here, TP (True Positives), FP (False Positives), and FN (False Negatives) denote the counts of true positives, false positives, and false negatives, respectively. 
True positives (TP) indicate correctly predicted occupied voxels, false positives (FP) indicate incorrectly predicted samples, and false negatives (FN) represent undetected occupied voxels.

Furthermore, the precision (P) of the prediction results can be derived. 
Precision indicates the proportion of correctly predicted samples among all samples predicted as occupied voxels. 
Higher precision implies that most of the regions predicted as occupied by the model are correct and can be calculated by:
\begin{equation} \centering
    \text{P} = \frac{\text{TP}}{\text{TP} + \text{FP}}
    \label{eq:precision}
\end{equation}

The recall (R) of the prediction results can also be calculated. 
Recall indicates the proportion of correctly predicted occupied regions among the actual occupied regions. 
Higher recall means the model can correctly predict most of the actually occupied regions as:
\begin{equation} \centering 
\text{R} = \frac{\text{TP}}{\text{TP} + \text{FN}}
\label{eq:recall}
\end{equation}

%%%%%%%%%%%%%%%%%%%%%%%%%%%%%%%%%%%%%%%%%%%%%%
\section{Results}
\label{sec:results}
%%%%%%%%%%%%%%%%%%%%%%%%%%%%%%%%%%%%%%%%%%%%%%

\begin{figure}[!ht]  \centering
    \includegraphics[width=0.48\textwidth]{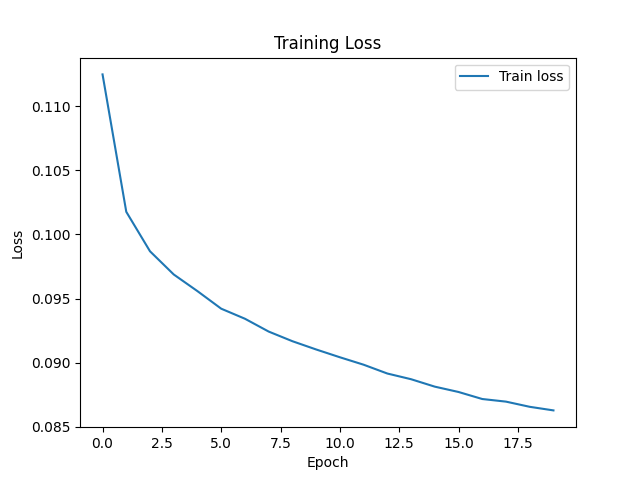}
    \caption{Loss curve for the first stage of training}
    \label{fig:loss_1}
\end{figure}

In this paper, we use a self-collected underground parking garage dataset as the training dataset. 
The dataset consists of 22 regions, with regions 0-10 used as the training set and regions 11-21 used as the test set.
The Loss curve for the first stage of training is shown in \autoref{fig:loss_1}.

\begin{figure}[!ht]  \centering
    \includegraphics[width=0.48\textwidth]{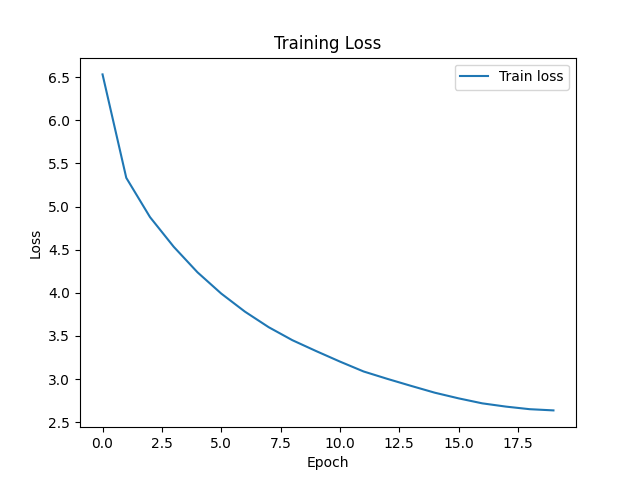}
    \caption{Loss curve for the second stage of training}
    \label{fig:loss_2}
\end{figure}

the second stage of training is shown in \autoref{fig:loss_2}.

The two figures above illustrate the change in binary cross-entropy loss during the training process over the epochs.
As the number of epochs increases, the training loss decreases, indicating that the model does not encounter the issue of vanishing gradients during the 20 epochs of training and continues to reduce the gap between the predicted values and the ground truth.
Despite the fluctuations observed in the loss curve during the first stage, the overall trend remains downward, suggesting that the model does not suffer from over-fitting.

\begin{figure}[!ht]  \centering
    \includegraphics[width=0.485\textwidth]{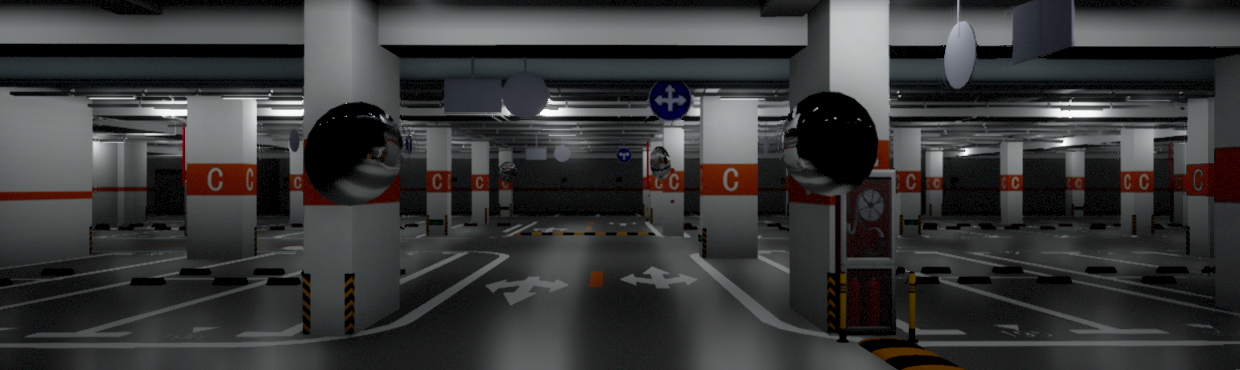}
    \caption{Input camera images}
    \label{fig:cam}
\end{figure}

The input camera images are shown in \autoref{fig:cam}.

\begin{figure}[!ht]  \centering
    \includegraphics[width=0.485\textwidth]{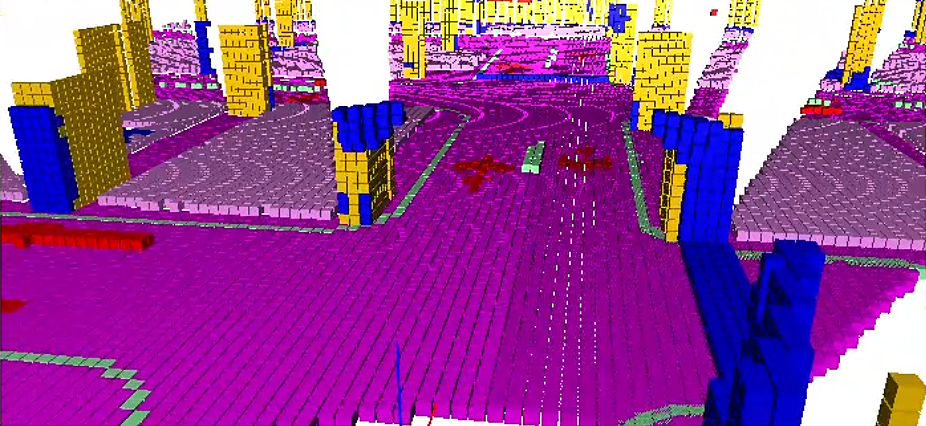}
    \caption{Model prediction results: color semantics correspond to \autoref{fig:uncompleted}.}
    \label{fig:pred}
\end{figure}

\begin{figure}[!ht] \centering
    \includegraphics[width=0.48\textwidth]{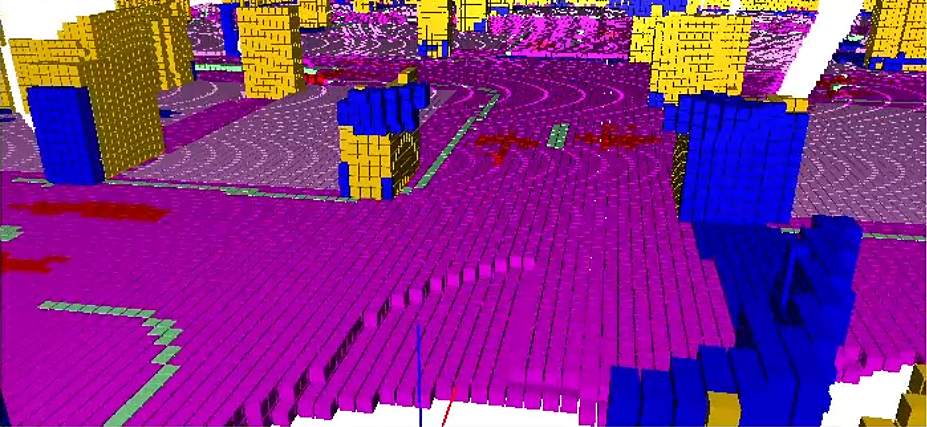}
    \caption{Ground truth: color semantics correspond to \autoref{fig:uncompleted}.}
    \label{fig:gt}
\end{figure}

The model prediction results and their comparison with the ground truth are shown in \autoref{fig:pred} and \autoref{fig:gt}.

The model's performance on the constructed dataset before and after training is shown in \autoref{tab:performance_metrics}.

\begin{table}[!ht]  \centering
    \caption{Performance metrics of the model before and after training}
    \label{tab:performance_metrics}
    \begin{tabular}{ccc} \toprule
        \textbf{Metric} & \textbf{Before Training} & \textbf{After Training} \\   \midrule
        IoU             & 37.08                    & 43.55                   \\
        Precision (P)   & 51.74                    & 61.04                   \\
        Recall (R)      & 54.82                    & 59.82                   \\
        mIoU            & 11.30                    & 13.25                   \\
        \bottomrule
    \end{tabular}
\end{table}

It can be observed that after training with the dataset collected in this paper, the model's performance is superior to that of the model trained solely on the KITTI dataset. 
Therefore, it can be concluded that the proposed approach enhances the perception performance of the occupancy grid network model in underground parking scenarios.

%%%%%%%%%%%%%%%%%%%%%%%%%%%%%%%%%%%%%%%%%%%%%%
\section{Discussion}
\label{sec:discussion}
%%%%%%%%%%%%%%%%%%%%%%%%%%%%%%%%%%%%%%%%%%%%%%

Due to limitations in time and personal ability, there are still areas in this paper that need improvement. 
For instance, the number of training epochs could be increased, given the constraints on computational power and scene modeling. 
Additionally, the diversity of the data collected needs enhancement. 
The excellent performance demonstrated may be due to the high similarity between the training and testing data.

Therefore, further work is required to explore the diversity of the dataset to enhance the model's generalization ability. 
This will ensure that the model can be widely applied in various underground parking lot scenarios and perform with the same perception capabilities as demonstrated in this experiment.

%%%%%%%%%%%%%%%%%%%%%%%%%%%%%%%%%%%%%%%%%%%%%%
\section{Conclusions}
\label{sec:conclusion}
%%%%%%%%%%%%%%%%%%%%%%%%%%%%%%%%%%%%%%%%%%%%%%

In this paper, we designed a data collection scheme based on the CARLA simulator environment, including semantic mapping, point cloud voxelization, and multi-frame fusion voxel completion algorithms. 
The completion algorithm enhances data reliability, providing stronger robustness.

The VoxFormer model, based on a powerful deformable attention mechanism and combining its depth prediction model with a downsampling occupancy grid prediction model, achieves efficient and accurate occupancy grid prediction. 
After 24 epochs of training, VoxFormer consistently achieved good prediction results in the underground parking lot scenario.

From the above results, it can be concluded that fine-tuning the model with the underground parking lot data collected in this paper allows it to adapt well to the lighting conditions and road conditions of the underground parking environment, exhibiting good perception performance.

\bibliographystyle{IEEEtran}
\bibliography{bibliography}

\end{document}